\documentclass[journal]{new-aiaa}

\usepackage{subcaption}
\usepackage{graphicx}
\usepackage{amsmath, bbm}
\usepackage[version=4]{mhchem}
\usepackage{siunitx}
\usepackage{longtable,tabularx}
\usepackage{breqn}
\usepackage{placeins}

\usepackage{hyperref}
\usepackage{cleveref}

\crefname{equation}{}{}

\usepackage{xcolor} 
\usepackage{colortbl}
\usepackage{color}

\newtheorem{helptheorem}{Theorem}{} 
 
\newtheorem{helplemma}[helptheorem]{Lemma} 
\newtheorem{helpcorollary}[helptheorem]{Corollary} 
 
\newtheorem{helpexample}[helptheorem]{Example} 
 
\newtheorem{helpproposition}[helptheorem]{Proposition} 
 
\newtheorem{helpremark}[helptheorem]{Remark} 
 
\newtheorem{helpdefinition}[helptheorem]{Definition}
 
\newtheorem{helpassumption}[helptheorem]{Assumption}

\newtheorem{helpproblem}[helptheorem]{Problem}

\begin{document}

\title{Close-Proximity Satellite Operations through Deep Reinforcement Learning and Terrestrial Testing Environments \footnote{The views expressed are those of the author and do not necessarily reflect the official policy or position of the Department of the Air Force, the Department of Defense, or the U.S. government. Approved for public release; distribution is unlimited. Public Affairs approval \#AFRL-2025-0336.}}
\author{Henry Lei\thanks{Software Controls Engineer, Verus Research 6100 Uptown Blvd NE, Albuquerque, New Mexico, 87110},
Joshua Aurand\thanks{Lead - Artificial Intelligence and Machine Learning},
Zachary S. Lippay\thanks{Lead - Dynamics and Controls, Verus Research},
\\ and Sean Phillips\thanks{Technology Advisor, Space Control Branch, Air Force Research Laboratory, Kirtland AFB, NM., 87117.}}

\date{January 2025}

\maketitle

\begin{abstract}
    With the increasingly congested and contested space environment, safe and effective satellite operation has become increasingly challenging. As a result, there is growing interest in autonomous satellite capabilities, with common machine learning techniques gaining attention for their potential to address complex decision-making in the space domain. However, the "black-box" nature of many of these methods results in difficulty understanding the model's input/output relationship and more specifically its sensitivity to environmental disturbances, sensor noise, and control intervention. This paper explores the use of Deep Reinforcement Learning (DRL) for satellite control in multi-agent inspection tasks. The Local Intelligent Network of Collaborative Satellites (LINCS) Lab is used to test the performance of these control algorithms across different environments, from simulations to real-world quadrotor UAV hardware, with a particular focus on understanding their behavior and potential degradation in performance when deployed beyond the training environment.
\end{abstract}

\section{Introduction}

Due to the increase in launch capability, access to the space domain has seen a reduction in cost. This has led to an increase in the number of active assets launched, creating an increasingly cluttered environment. This makes ground operation more complex where safe vehicle operation is becoming increasingly difficult to guarantee. To help manage the difficulty in scaling ground operation to meet the increasing number of vehicles in orbit, autonomous satellite capabilities have been increasingly sought after in the design of safe space missions. To exemplify the shift towards autonomous satellites there has been significant attention in developing itemized levels of autonomous satellite operations \cite{10115976,baker2025,ANTSAKLIS202015}.  
Many tools in control theoretic fields have already been leveraged to provide operating satellites with autonomous capability, but machine learning and cognitive artificial intelligence (AI), in particular, have garnered much interest - as well as concerns - in the community due to their demonstrated potential capabilities in solving difficult decision making problems in a variety of domains; for example in space operation see \cite{JSR24, InfoInspection, vanWijk23,Lei22,Bernhard20,Nakka21}. 

With on-orbit navigation and planning tasks being largely handled through the optimal control of highly nonlinear, strongly coupled dynamical systems, a variety of simplifying assumptions are imposed on planning scope, application, or capability. For instance, \cite{Maestrini22,oestreich21,Dor18} focus on an autonomous multi-vehicle inspection mission used to help develop controllers for effective close proximity satellite operation. To help address this, tools in machine learning are being developed leveraging DRL for guidance, navigation, and control; see \cite{JSR24,InfoInspection,vanWijk23,Lei22}. However, there are a couple of general factors that still limit the widespread adoption of RL-based policies by decision-makers. For one, as a black-box method, it is difficult to guarantee or even understand the decision-making process. Small disturbances, even as minor as a single pixel in a large image (\cite{Machado21,Pitropakis19}), can cause large instabilities in the decision-making - a well-known issue in the field of robust reinforcement learning (RL) \cite{moos_robust_2022}. In the case of deploying the trained policy into a new environment (eg. from a simulated environment to hardware-in-the-loop), the cumulative differences can significantly degrade the performance of the trained policy. Bridging the so-called \textit{sim2real} gap (see \cite{pan2010survey} is the study of a subset of the field of transfer learning and domain adaptation \cite{zhuang_comprehensive_2020,tobin2017domain}. 

The topic of robustness itself has been extensively investigated in optimization theory; therein an underlying uncertainty set is typically specified - under which the optimizer must manage all possible deviations induced by such uncertainty. Since the RL agents continuously observes system response through interaction, their ability to effectively sample low-probability outcomes within the uncertainty set is limited. This difficulty is exaggerated when access to real and/or representative system data is limited (such as in the space environment). More details regarding this can be found in~\cite{morimoto2005, pinto2017, tamar2013, tessler2019}. Although such difficulties exist, there are some key advantages to training on simulated data. For instance, this commonly allows for accelerated data collection through parallelization, safe experimentation with failure scenarios, and access to extreme conditions that are impractical or impossible to replicate in the real world---such as those encountered during long-duration space missions~\cite{kalakrishnan2013learning, nguyen2020deep}. To take advantage of this in a meaningful way, it is crucial to analyze how the byproduct (a trained RL control policy) will perform in real-world environments beyond the scope implied by simulation. To do so, we leverage the LINCS lab, which offers a tested platform for addressing such challenges. By utilizing a terrestrial testbed with quadrotors emulating satellite on-orbit dynamics, LINCS enables detailed evaluations of policy performance under controlled experimental conditions. This facility provides the infrastructure and hardware necessary for collecting trajectory and performance data across simulated and physical scenarios~\cite{phillips2024emulation}. Controller robustness and performance degradation will be tested in three distinct environments: (1) the training environment, (2) the LINCS simulation platform, and (3) through quadrotor UAV hardware emulating the space environment. 

The work presented below is concurrently explored in a companion paper, see \cite{Lei25}. Where \cite{Lei25} focuses on the robustness of DRL based \textit{guidance generation} when measured against changes in \textit{mission performance}, this work explores the impact that intentional and significant control intervention can have on multi-agent low-level DRL based \textit{control}. This is done by constructing experiments resulting in a high likelihood of RTA intervention for constraint violation including agent-to-agent distance and vehicle speed. A brief background covering key notation, selected results from \cite{Lei25}, problem formulation, and the LINCS Lab is provided in \textit{Background}.  This is followed by our \textit{Experimental Setup}, key results in \textit{Results}, and concluding remarks in \textit{Conclusions}.

\section{Background}\label{sec: Background}
\subsection{Commonly Used Notation}
We adopt notation from \cite{Lei25}. Therein, the Earth-Centered Inertial (ECI) frame, denoted as $F_E$, is defined by the orthogonal unit vectors $\mathbf{i}_E$, $\mathbf{j}_E$, and $\mathbf{k}_E$, with origin at $O_E$. The $\mathbf{i}_E$ direction points toward the vernal equinox, $\mathbf{k}_E$ aligns with Earth's north celestial pole, and $\mathbf{j}_E$ completes the right-handed coordinate system. For $n \in \mathbb{N}$ deputy spacecraft indexed by $\mathcal{I}_d = \{1, 2, \dots, n\}$ and a chief spacecraft (alternatively referred to as an RSO) indexed as $i = 0$, the full agent set is $\mathcal{I} = \mathcal{I}_d \cup \{0\}$. Each spacecraft $i \in \mathcal{I}$ has a body-fixed frame $F_i$, defined by orthogonal unit vectors $\mathbf{i}_i$, $\mathbf{j}_i$, and $\mathbf{k}_i$, with origin at $O_i$. The Hill's frame, $F_H$, is fixed to the chief spacecraft, with origin $O_H = O_0$. Its axes are defined as follows: $\mathbf{i}_H$ points radially from $O_0$ toward $O_E$, $\mathbf{k}_H$ aligns with the chief's angular momentum vector, and $\mathbf{j}_H$ completes the right-handed system. The position of spacecraft $i \in \mathcal{I}$ relative to $O_E$ is denoted by $\mathbf{r}_i \in \mathbb{R}^3$, $\mu$ denotes Earth's gravitational parameter, and $m_i$ the spacecraft mass. Dynamics perturbation force due to Earth's oblateness is described by the $J_2$ coefficient. The relative position vector with respect to the chief spacecraft is defined as $\delta\mathbf{r}_i = \mathbf{r}_i - \mathbf{r}_0$. Resolved in Hill's frame $F_H$, $\delta\mathbf{r}_i \in \mathbb{R}^3$ describes the deputy’s relative motion with respect to the chief.

\subsection{LINCS Lab}
\begin{figure}[!htb]
\begin{subfigure}{0.45\textwidth}
\includegraphics[width=0.99\linewidth]{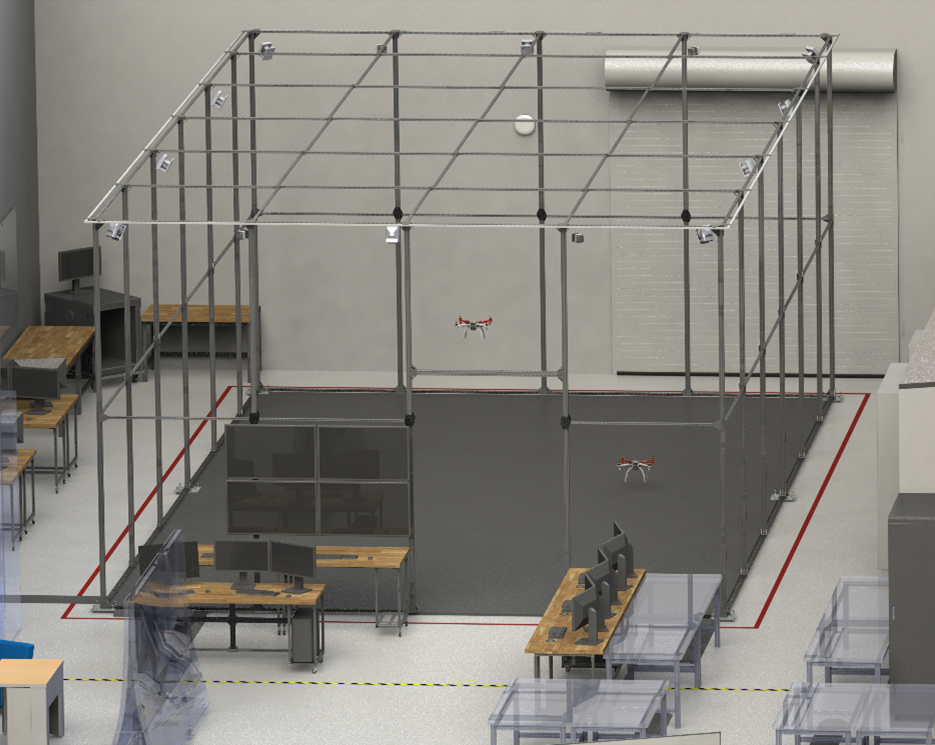} 
\label{fig:Aviary_CAD}
\caption{CAD model of the Aviary testbed facility.}
\end{subfigure}
\begin{subfigure}{0.5\textwidth}
\includegraphics[width=0.99\linewidth]{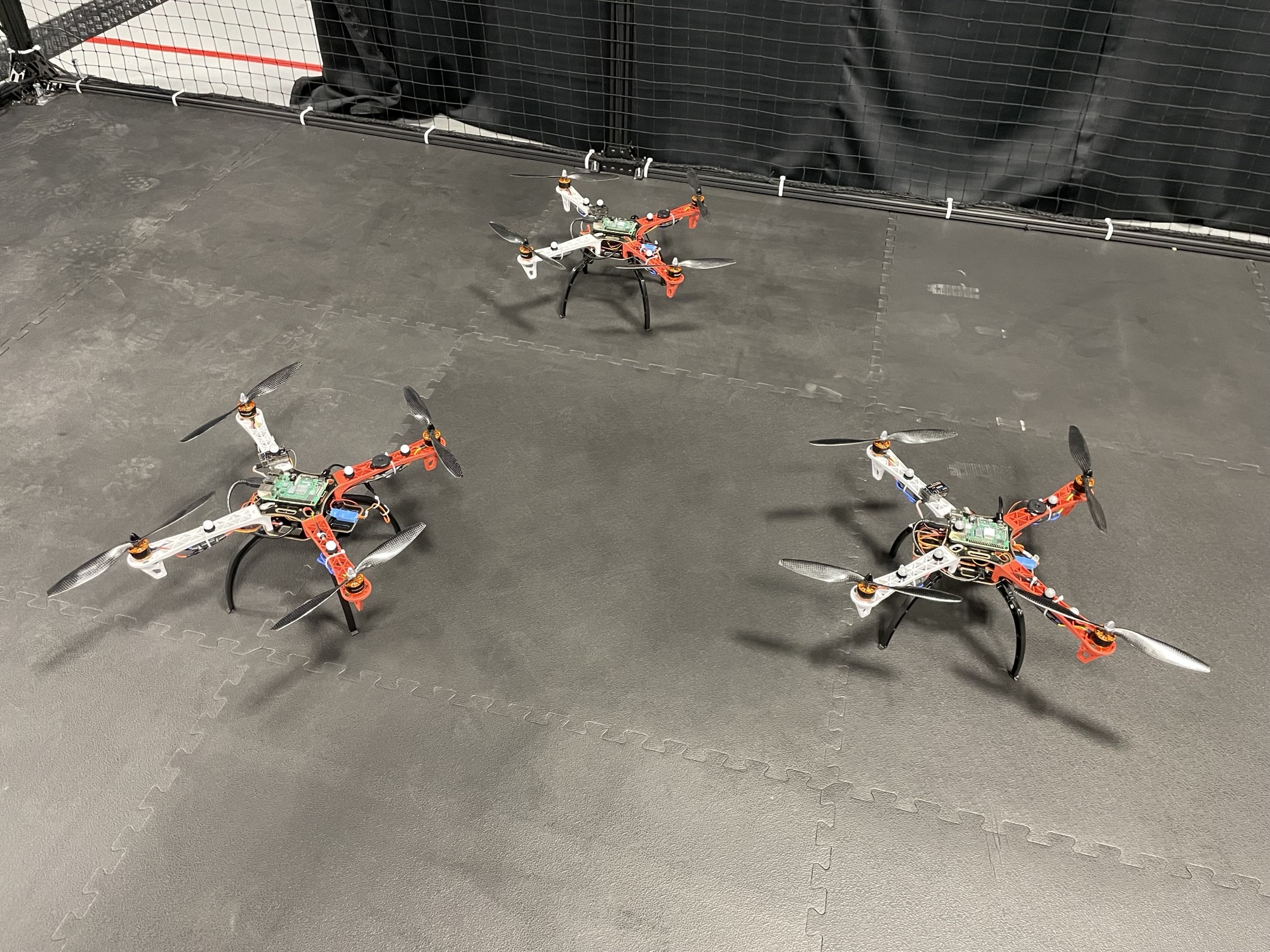}
\label{fig:Aviary_actual}
\caption{Squadron of quadcopters currently utilized in the lab.}
\end{subfigure}
\caption{Local Intelligent Network of Collaborative Satellites (LINCS) Laboratory overview}
\label{fig:Aviary}
\end{figure}

To emulate the close proximity interactions between multiple vehicles, we leverage the 
Local Intelligent Networked Collaborative Satellites (LINCS) laboratory. This facility contains a simulation and emulation environment designed for testing rendezvous and proximity operations of multi-vehicle satellite guidance, navigation, control, and autonomy algorithms \cite{phillips2024emulation}. To offset gravity interactions the lab utilizes quadcopters to emulate spacecraft motion inside a bounded capture space. We represent the relative space frame called the Hill's frame with the origin as the center of the capture space; for example see \cref{fig:Aviary}. The lab environment utilizes \textit{Robot Operating System 2 (ROS2)} to pass messages between different dynamics, plant, and flight software nodes. This structure is similar to how an algorithm would be tested if it were running onboard a spacecraft. 

The guidance, navigation, and control algorithms are simulated using 2-body dynamics with J2 gravitational perturbations, and these trajectories are emulated using the quadcopters in the physical lab space such that they appear to follow the space trajectories. Importantly, these trajectories must be scaled from the space frame to the lab frame so they fit within the dimensions of the lab space. A Vicon Motion Capture System is used to determine the physical state of the quadcopters.

\begin{figure}[!htb]
\begin{subfigure}[b]{.45\textwidth}
    \centering
    \includegraphics[width=\textwidth]{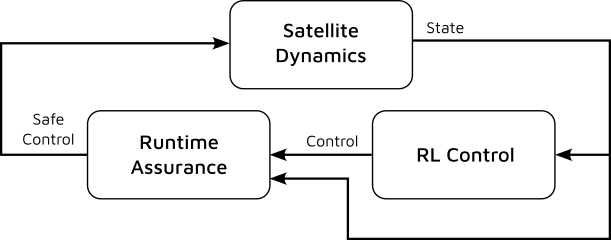}
\end{subfigure}
\hfill
\begin{subfigure}[b]{0.45\textwidth}
\centering
\includegraphics[width=\textwidth]{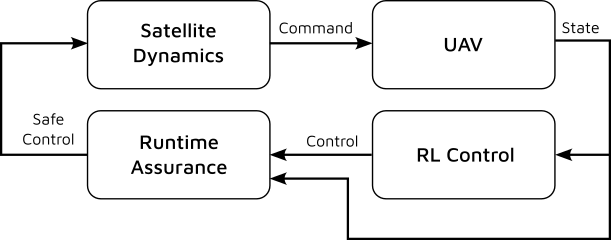}
\end{subfigure}
\caption{The left diagram shows simulation workflow with RTA enforcement in the LINCS Lab. The right diagram shows an analogous version for LINCS Cyber-Physical emulation using quadrotor UAVs.}
\label{fig:LINCS_block_diagram}
\end{figure}

To ensure safety in the lab, a lab safety RTA filter is applied to the control of each quadcopter. This RTA is based on quadcopters dynamics, rather than space dynamics, and is designed to prevent the quadcopters from colliding with each other or the walls of the lab, as well as to limit the acceleration along each axis. This allows unverified algorithms to be tested in the lab without the risk of damaging the quadcopters or the lab space. Further descriptions of the LINCS lab and quadcopters RTA algorithms are beyond the scope of this paper, but are provided in \cite{phillips2024emulation}. Block diagrams for UAV in the loop and out of the loop can be seen in \cref{fig:LINCS_block_diagram}.

\subsection{Reference Problem}

\begin{figure}[!htb]
    \includegraphics[width=\textwidth]{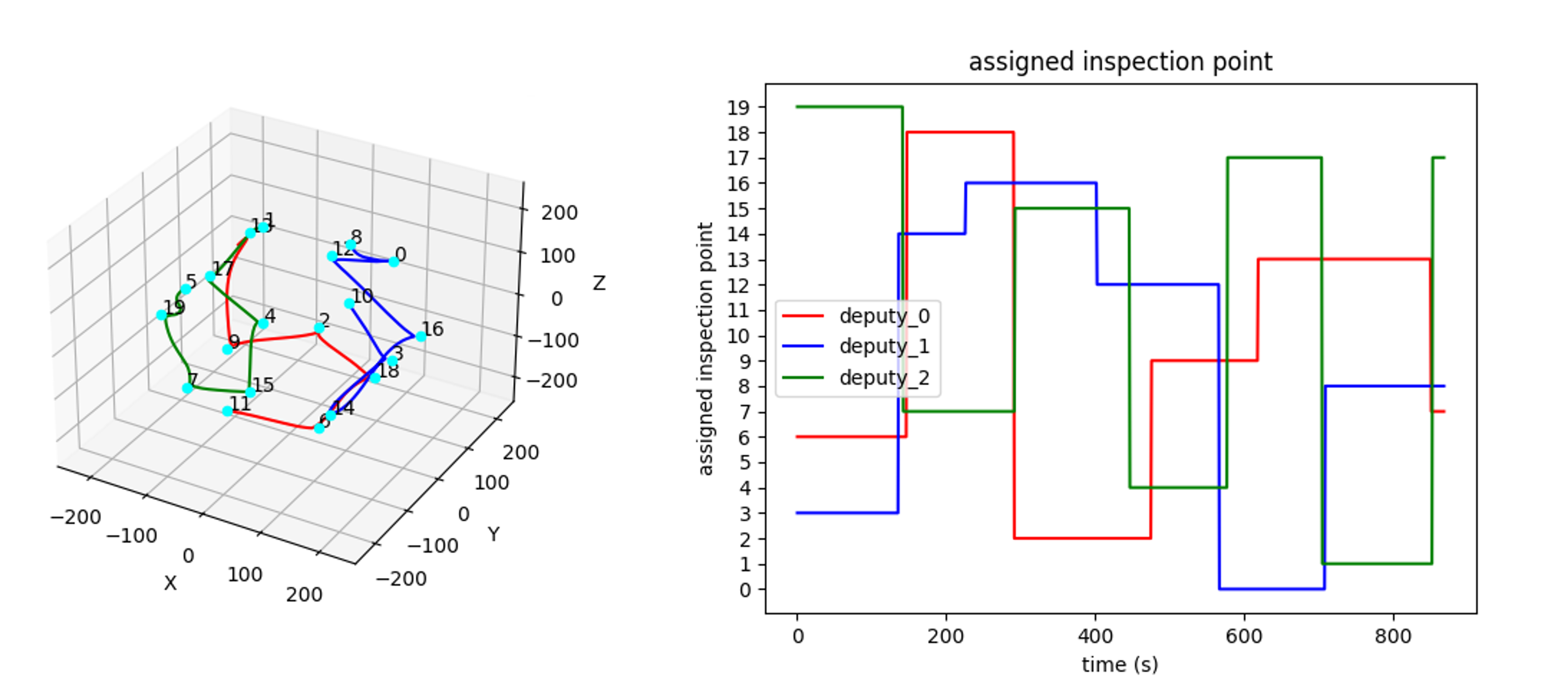}
    \caption{The left diagram shows sample trajectories generated using the DRL hierarchical controller proposed in \cite{Lei22}. The right diagram details task assignment by the HL planner. The solution presented is formulated to provide visually diverse viewing conditions to collaboratively inspect an unknown RSO, also called ``Chief".}
    \label{fig:HierControl}
\end{figure}

\subsubsection{Hierarchical Control} 

In \cite{Lei25}, a hierarchical inspection controller was deployed in a simulated environment to evaluate its ability to integrate high-level guidance and low-level point-to-point control policies. Based on the formulation in \cite{Lei22}, a series of experiments were then conducted to assess the controller's effectiveness in achieving full coverage of inspection targets while maintaining task efficiency and stability under modeled trajectory uncertainties. The environment consisted of 20 inspection points arranged in a graph-based ellipsoid, with three agents collaboratively performing the inspection task. Key metrics such as graph coverage, time taken, distance traveled, and fuel consumption were measured and compared to a statically defined performance baselines. This was constructed to assess the impact that policy inference timing asynchronicity has on task efficacy. The hierarchical controller itself was composed of a pretrained guidance planner (High-level planning) and point-to-point motion controller (Low-level control). The results presented in Table 5 of \cite{Lei25} demonstrate that the hierarchical control policy was capable of cointegrating guidance and control policies to achieve robust multi-agent coordination. Although performance degraded with exposure to different testing environments, task performance for graph coverage and distance traveled were not significantly impacted. By progressively increasing fidelity and complexity, the experiments highlighted the framework's ability to bridge the sim-to-real gap and adapt to real-world disturbances. These include sensor feedback, dynamic perturbations, and minor RTA intervention. Overall, this study provides evidence of the hierarchical DRL framework's potential for enabling robust and efficient autonomous satellite operations in increasingly complex and dynamic environments.

\subsubsection{Low-level Control}\label{subsec: LL control}
While \cite{Lei25} investigated stability of the DRL hierarchical control scheme presented in \cite{Lei22} under nominal operating conditions, this work focuses on assessing the impact of \textit{low-likelihood scenarios} on controller performance. To do this, we focus on the careful control of HL planning targets used to increase the relative likelihood of improbable scenarios for the LL control scheme. This allows for the direct exploration of performance in cases that were too infrequently sampled for the analysis of \cite{Lei25} to adequately encapsulate. For background, the LL control policy attempts to solve a motion control problem modeling the constrained movement of agents between assigned waypoints. For an arbitrary agent, the motion control problem is defined through the calculation of:

\begin{equation}\label{eqn: LL_control}
    \mathbf{u}^* = \arg \max_{\mathbf{u}(t)} \int_0^T \frac{1}{\|\delta\mathbf{r}(t) - \delta\mathbf{r}_{g}\| + 1} \, dt,
\end{equation}
\hspace{20ex}subject to:
\begin{equation}
    \delta\mathbf{r}(T) = \delta\mathbf{r}_g,\,\delta\mathbf{r}(t) = \mathbf{f}_{\text{Hill}}(t,\mathbf{u}(t)),\text{ and }\|\mathbf{u}(t)\|_{\infty} \leq 1\quad  \forall t \in [0, T].
\end{equation}

\begin{figure}[!htb]
\begin{subfigure}[b]{.5\textwidth}
    \centering
    \includegraphics[width=\textwidth]{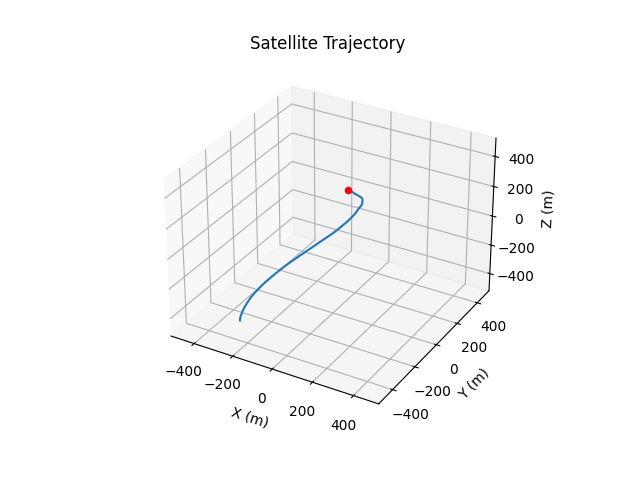}
\end{subfigure}
\hfill
\begin{subfigure}[b]{0.5\textwidth}
\centering
\includegraphics[width=\textwidth]{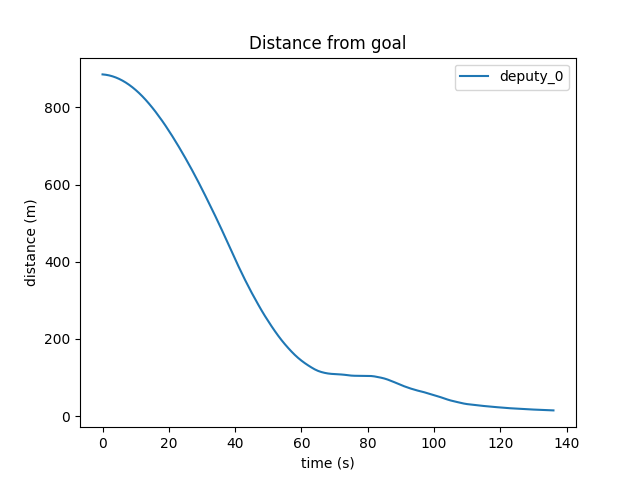}
\end{subfigure}
\caption{The left diagram shows a single agent satellite trajectory driven by the DRL LL controller from \cite{Lei22}. The right diagram shows distance to goal over time; an episode is considered successful once the satellite reaches a fixed threshold around the goal waypoint.}
\label{fig:LL Controller}
\end{figure}

The objective function in \cref{eqn: LL_control} incentivizes movement that results in distance change inversely proportional to distance to the target waypoint, encouraging proximity to the goal. Constraints include: a terminal condition requiring the vehicle to reach a goal location $\delta\mathbf{r}_{g}$, translational dynamics are updated using linearized relative motion $\mathbf{f}_{\text{Hill}}(t,\mathbf{u}(t))$ (see Section II-D of \cite{Lei25}), and uniformly bounded axially aligned thrust inputs. This problem is solved using the DRL based formulation from \cite{Lei22}; the environment specification described in \cite{Lei25} is borrowed to solve \cref{eqn: LL_control}. Practically, at each discrete time step (\(\Delta t = 1\) s), the trainable agent receives a state observation containing distance and direction of $\delta \mathbf{r}_g$ to the deputy in conjunction with a relative velocity measurement. This can be seen below:
\begin{equation}\label{eqn:llobs}
    \mathbf{O}_{ll} = \left( \frac{\delta \mathbf{r} - \delta \mathbf{r}_g}{1000}, \delta \dot{\mathbf{r}} \right).
\end{equation}
The episode terminates under the following conditions:
\begin{enumerate}
    \item The agent successfully reaches the target, defined as \(\|\delta \mathbf{r} - \delta \mathbf{r}_g\| < 10 \, \text{m}\),
    \item The agent drifts out of bounds,
    \item The episode duration exceeds 500 s.
\end{enumerate}
Training instantiation is rescaled according to dimensions provided by the LINCS aviary where each instance is initialized with a randomized agent state drawn according to 
\begin{equation}\label{eqn:sampling}
    \delta \mathbf{r}(0) = [1.17, 2.5, 1]\cdot\mathbf{x}, \quad \delta \mathbf{r}_g = [1.17, 2.5, 1]\cdot \mathbf{y},
\end{equation}
where $\mathbf{x}, \mathbf{y} \sim \mathcal{U}([-240\text{m}, 240\text{m}]^3)$. The reward function used for training LL control is designed to minimize distance to the target waypoint $\delta \mathbf{r}_g$ in conjunction with the objective in eqn.~\eqref{eqn: LL_control}. The function is conditioned on positional state $\delta \mathbf{r}_{-1}$ and an implicitly defined thrust action $\mathbf{u}$. The effect of $\mathbf{u}$ on the reward $\textrm{rew}(\cdot)$ is encoded by the terms $\delta \mathbf{r}$ and $\delta \dot{\mathbf{r}}$ reflecting vehicle state transition under $\mathbf{u}$ for a single time step. Inspired by the Langrangian associated with eqn.~\eqref{eqn: LL_control}, the deputy is rewarded at each time step based on the following:
\begin{align}
    \textrm{rew}(\delta \mathbf{r}, &\delta \mathbf{r}_{-1}, \delta \dot{\mathbf{r}}, \delta \mathbf{r}_g) =\notag\\ 
    &\frac{\alpha}{\|\delta \mathbf{r} - \delta \mathbf{r}_g\| + 1} 
    + \beta \left( \|\delta \mathbf{r}_{-1} - \delta \mathbf{r}_g\| - \|\delta \mathbf{r} - \delta \mathbf{r}_g\| \right) 
    - \nu \|\delta \dot{\mathbf{r}}\|_1 \mathbb{I}_{\{\|\delta \dot{\mathbf{r}}\| > \eta \sigma_\mu \|\delta \mathbf{r} - \delta \mathbf{r}_g\| \}}\label{eqn:rew_ll}
\end{align}
where $\mathbb{I}_{\{\cdot\}}$ is the indicator function, \(\alpha, \beta, \nu\) are weighting factors, and \(\sigma_\mu\) defines the maximum allowable velocity threshold. The first and second terms incentivize distance reduction to a goal waypoint with additional reinforcement provided to movement along a straight line. The third term encodes a variable speed limit penalizing conditions that may be unsafe or difficult to control. An example LL control trajectory can be seen below in \cref{fig:LL Controller}.

\subsection{Runtime Assurance}
To enforce satellite safety during testing, we leverage a constrained minimum intervention quadratic program imposing constraints on satellite interagent distance, velocity, and acceleration.

\underline{Interagent Distance:} Restrictions on inter-agent distance ensure collision free operation while in proximity of other satellites and the chief object. For all $(i,j) \in \mathcal{P} \triangleq \{ (i,j) \in \mathcal{I}_{\rm d} \times \mathcal{I}_{\rm d} : i \neq j \}$, ie. the set of all interagent pairs, define:
\begin{equation} \label{pos:bf:h}
    h_{ij}\left( \delta \mathbf{r}_i, \delta \mathbf{r}_j \right) = \frac{1}{2}\left( \left( \delta \mathbf{r}_i  - \delta \mathbf{r}_j  \right)^{\rm T} \left( \delta  r_i  - \delta \mathbf{r}_j  \right) - r_c^2 \right),
\end{equation}
where $r_c > 0$ is a predefined constant describing collision radius. We assume the relative positions and velocities of each agent are known, ensuring consensus on $O_{H}$ as expressed in $F_{E}$. It follows from \cref{pos:bf:h} by taking the partial derivative of $h_r$ along the trajectories of $\delta \mathbf{r}_i, \delta \mathbf{r}_j$ that
\begin{align}
    \dot{h}_{ij} \left( \delta \mathbf{r}_i, \delta \mathbf{r}_j, \delta \dot{\mathbf{r}}_i, \delta \dot{\mathbf{r}}_j \right) &= \frac{\partial h_{ij}}{\partial \delta \mathbf{r}_i} \delta \dot{\mathbf{r}}_i +  \frac{\partial h_{ij}}{\partial \delta \mathbf{r}_j} \delta \dot{\mathbf{r}}_j = \left( \delta \mathbf{r}_i - \delta \mathbf{r}_j \right)^{\rm T} \left( \delta  \dot{r}_i - \delta \dot{\mathbf{r}}_j  \right). \label{pos:bf:hdot}
\end{align}
Taking the derivative of $\dot{h}_{ij}$ along the trajectories $\delta \mathbf{r}_i, \delta \mathbf{r}_j, \delta \dot{\mathbf{r}}_i, \delta \dot{\mathbf{r}}_j$ then gives:
\begin{align*}
    &\ddot{h}_{ij} \left( \delta \mathbf{r}_i, \delta \mathbf{r}_j, \delta \dot{\mathbf{r}}_i, \delta \dot{\mathbf{r}}_j, \delta \ddot{\mathbf{r}}_j, u_i \right) = \frac{\partial \dot{h}_{ij}}{\partial \delta \mathbf{r}_i} \delta \dot{\mathbf{r}}_i +  \frac{\partial \dot{h}_{ij}}{\partial \delta \mathbf{r}_j} \delta \dot{\mathbf{r}}_j + \frac{\partial \dot{h}_{ij}}{\partial \delta \dot{\mathbf{r}}_i} \delta \ddot{\mathbf{r}}_i +  \frac{\partial \dot{h}_{ij}}{\partial \delta \dot{\mathbf{r}}_j} \delta \ddot{\mathbf{r}}_j 
\end{align*}
Simplifying the expression and substituting relative motion dynamics defined in eqn.~1 of \cite{JSR24}, this is equivalently expressed as: 
\begin{subequations} \label{pos:bf:hddot}
\begin{align}
    &\ddot{h}_{ij} \left( \delta \mathbf{r}_i, \delta \mathbf{r}_j, \delta \dot{\mathbf{r}}_i, \delta \dot{\mathbf{r}}_j, \delta \ddot{\mathbf{r}}_j, u_i \right) = \notag \\
    & || \delta \dot{\mathbf{r}}_i - \delta \dot{\mathbf{r}}_j ||^2 - \left( \delta \mathbf{r}_i - \delta \mathbf{r}_j \right)^{\rm T} \delta  \ddot{r}_j +  \left( \delta \mathbf{r}_i  - \delta \mathbf{r}_j  \right)^{\rm T} \bigg( f_i \left( \delta \mathbf{r}_i, \delta \dot{\mathbf{r}}_i \right) + g_i(\delta \mathbf{r}_i, \delta \dot{\mathbf{r}}_i) u_i \bigg) \label{pos:bf:hddot:1} \\
    & \text{where:} \notag \\
    & \hspace{2ex} f_i(\delta \mathbf{r}_i, \delta \dot{\mathbf{r}}_i) = \begin{bmatrix}
        3 \sigma_{\mu}^2 & 0 & 0 & 0 & 2 \sigma_\mu & 0 \\
        0 & 0 & 0 & -2 \sigma_{\mu} & 0 & 0 \\
        0 & 0 & -\sigma_{\mu}^2 & 0 & 0 & 0 
    \end{bmatrix} \begin{bmatrix}  \delta \mathbf{r}_i \\ \delta \dot{\mathbf{r}}_i  \end{bmatrix}\text{ and }
    g_i(\delta \mathbf{r}_i, \delta \dot{\mathbf{r}}_i) = \frac{1}{m_{i}}\mathbf{I}_{3\times3}\label{pos:bf:hddot:2}.
\end{align}
\end{subequations}

This is used to formulate a collision avoidance constraint utilizing higher-order barrier functions. More specifically, let $\gamma_0 >0$ and $\gamma_1 >0$ be fixed constants. Suppose the following constraints hold for all $t \geq 0$:
\begin{align}
   &h_{ij} \left( \delta \mathbf{r}_i(0), \delta \mathbf{r}_j(0) \right) >0,\\
   &\dot{h}_{ij} \left( \delta \mathbf{r}_i (0), \delta \mathbf{r}_j(0), \delta \dot{\mathbf{r}}_i(0), \delta \dot{\mathbf{r}}_j(0) \right) + \gamma_0 h_{ij} \left( \delta \mathbf{r}_i(0), \delta \mathbf{r}_j(0) \right) > 0,\\
   &\ddot{h}_{ij} \left( \delta \mathbf{r}_i (t), \delta \mathbf{r}_j (t), \delta \dot{\mathbf{r}}_i (t), \delta \dot{\mathbf{r}}_j (t), \delta \ddot{\mathbf{r}}_j (t), u_i (t) \right) \notag \\ 
    & \hspace{10ex} + \left(\gamma_1 + \gamma_0 \right) \dot{h}_{ij} \left( \delta \mathbf{r}_i (t), \delta \mathbf{r}_j (t), \delta \dot{\mathbf{r}}_i (t), \delta \dot{\mathbf{r}}_j (t) \right) + \gamma_1 \gamma_0 h_{ij} \left( \delta \mathbf{r}_i(t), \delta \mathbf{r}_j(t) \right) \geq 0.\label{pos:constraint}
\end{align}
Following standard arguments in \cite{hocbf_Xiao_22}, for all $t \geq 0$:
\begin{align*}
    &h_{ij} \left( \delta \mathbf{r}_i(t), \delta \mathbf{r}_j(t) \right) \geq 0,\\
    &\dot{h}_{ij} \left( \delta \mathbf{r}_i (t), \delta \mathbf{r}_j(t), \delta \dot{\mathbf{r}}_i(t), \delta \dot{\mathbf{r}}_j(t) \right) + \gamma_0 h_{ij} \left( \delta \mathbf{r}_i(t), \delta \mathbf{r}_j(t) \right) \geq 0.
\end{align*}
As such, if agents are instantiated in a safe state (positionally) and an admissible (eg. satisfying \cref{pos:constraint}) control $u_i$ exists, then agent $i$ is able to maintain a minimum distance $r_c$ away from agent $j$. Without loss of generality, this applies to the chief if $\delta \mathbf{r}_j (t) = \delta \dot{\mathbf{r}}_j (t) = \delta \ddot{\mathbf{r}}_j (t) = 0$ for all $t \geq 0$ in \cref{pos:constraint}.

\underline{Velocity Constraint}
Velocity constraints ensure that vehicles move at a safe and controllable speed throughout operation. Consider
\begin{equation} \label{vel:bf:b}
    b_{i} \left( \delta \dot{\mathbf{r}}_i \right) = \frac{1}{2}\left( v_c^2 - \delta \dot{\mathbf{r}}_i ^{\rm T}  \delta  \dot{r}_i  \right),
\end{equation}
where $v_c > 0$ is the magnitude of a predetermined upper bound for relative velocity. It follows from \cref{vel:bf:b} that the derivative along the trajectory of $\delta \dot{\mathbf{r}}_i$ is
\begin{equation} \label{vel:bf:bdot}
    \dot{b}_i \left( \delta \mathbf{r}_i, \delta \dot{\mathbf{r}}_i, u_i \right) = \frac{\partial b_i}{ \partial \delta \dot{\mathbf{r}}_i  } \delta \ddot{\mathbf{r}}_i = - \delta \dot{\mathbf{r}}_i^{\rm T} \bigg( f_i \left( \delta \mathbf{r}_i, \delta \dot{\mathbf{r}}_i \right) + g_i(\delta \mathbf{r}_i, \delta \dot{\mathbf{r}}_i) u_i \bigg),
\end{equation}
where $f_i$ and $g_i$ are described in \cref{pos:bf:hddot:2}. Fix a constant $\gamma_2 >0$ and suppose for all $t \geq 0$ that the following constraints hold:
\begin{align}
        b_i \left( \delta \dot{\mathbf{r}}_i(0) \right) >0,\text{ and }
        \dot{b}_i \left( \delta \mathbf{r}_i(t), \delta \dot{\mathbf{r}}_i (t), u_i(t) \right) + \gamma_2 \dot{b}_i \left( \delta \dot{\mathbf{r}}_i (t) \right) \geq 0. \label{vel:constraint}
\end{align}
Following standard arguments in \cite{hocbf_Xiao_22}, for all $t \geq 0$, $b_i(\delta \dot{\mathbf{r}}_i) \geq 0$. As such, if agents are instantiated at a safe speed and an admissible (eg. satisfying \cref{vel:constraint}) control $u_i$ exists, then agent $i$ is able to maintain a safe speed.

\underline{Acceleration and Input Constraints}
Next we present constraints imposed on the acceleration and thrust input for the $i$-th agent. With regards to acceleration, we require:
\begin{align} \label{acc:constraint}
    a_c^2 - \delta \ddot{\mathbf{r}}_i^{\rm T} \bigg( f_i \left( \delta \mathbf{r}_i, \delta \dot{\mathbf{r}}_i \right) + g_i(\delta \mathbf{r}_i, \delta \dot{\mathbf{r}}_i) u_i \bigg) \geq 0,
\end{align}
where $f_i$ and $g_i$ are described in \cref{pos:bf:hddot:2}, and $a_c>0$ is a predetermined upper bound on the magnitude of vehicle acceleration. Note that $\delta \ddot{\mathbf{r}}_i$ represents the actual acceleration (e.g., measured or estimated) of the $i$-th agent. With regards to thrust input constraints, we require: 
\begin{equation} \label{input:constraint}
    \begin{bmatrix}
        (f_{c})_{(1)} \\ (f_{c})_{(2)} \\ (f_{c})_{(3)}
    \end{bmatrix} - 
    \begin{bmatrix}
        | (u_i)_{(1)}| \\ | (u_i)_{(2)}| \\ | (u_i)_{(3)}|  
    \end{bmatrix} \geq 0,
\end{equation}
where $(f_{c})_{(1)}, (f_{c})_{(2)}, (f_{c})_{(3)} >0$ are the upper bounds on the input $u_i$ to eqn.~1 of \cite{JSR24}. Note that \cref{input:constraint} is a \textit{box constraint} which is why each element of the input has its own upper bound.

\subsubsection{Relaxed Quadratic Program}
The constraints described above are used to define a quadratic program (QP) used to formalize the RTA objective. More concretely, fix any $i \in \mathcal{I}_{\rm d}$ and $(i,j) \in \mathcal{P}$. Introduce the slack variables for each agent $\phi_i := [\phi_{ij}, \phi_i^{(v)}, \phi_i^{(a)}, \phi_i^{(u,1)},\phi_i^{(u,2)},\phi_i^{(u,3)}]$ and consider a QP defined by:
\begin{subequations}
\begin{align}
    &(u_i, \phi_i) = \underset{u^*, \phi^* }{\text{arg min}} ||u^* - a_i ||^2 + \gamma_3 \phi^{* {\rm T}} \|\phi^*\|^2, \label{qp} \\
    & \hspace{-5ex} \text{subject to:} \notag \\
    & \hspace{-2ex} \ddot{h}_{ij} \left( \delta \mathbf{r}_i , \delta \mathbf{r}_j , \delta \dot{\mathbf{r}}_i , \delta \dot{\mathbf{r}}_j , \delta \ddot{\mathbf{r}}_j , u^* \right) \notag \\ 
    & + \left(\gamma_1 + \gamma_0 \right) \dot{h}_{ij} \left( \delta \mathbf{r}_i , \delta \mathbf{r}_j , \delta \dot{\mathbf{r}}_i , \delta \dot{\mathbf{r}}_j  \right) + \gamma_1 \gamma_0 h_{ij} \left( \delta \mathbf{r}_i, \delta \mathbf{r}_j \right) \geq \phi_{ij} \\
    & \hspace{-2ex} \dot{b}_i \left( \delta \mathbf{r}_i, \delta \dot{\mathbf{r}}_i , u^* \right) + \gamma_2 \dot{b}_i \left( \delta \dot{\mathbf{r}}_i  \right) \geq \phi_i^{(v)} \\
    & \hspace{-2ex} a_c^2 - \delta \ddot{\mathbf{r}}_i^{\rm T} \bigg( f_i \left( \delta \mathbf{r}_i, \delta \dot{\mathbf{r}}_i \right) + g_i(\delta \mathbf{r}_i, \delta \dot{\mathbf{r}}_i) u^* \bigg) \geq \phi_i^{(a)} \\
        & \hspace{-2ex} \begin{bmatrix}
        (f_{c})_{(1)} \\ (f_{c})_{(2)} \\ (f_{c})_{(3)}
    \end{bmatrix} - 
    \begin{bmatrix}
        | (u^*)_{(1)}| \\ | (u^*)_{(2)}| \\ | (u^*)_{(3)}|  
    \end{bmatrix} \geq \begin{bmatrix}
         \phi_i^{(u,1)} \\  \phi_i^{(u,2)} \\  \phi_i^{(u,3)}
    \end{bmatrix},
\end{align}
\end{subequations}
where $\gamma_3 >> 0$ is used as a penalty variable to ensure that $\phi_i$ remains as small as possible. Note that the slack variables are used to ensure feasibility of the optimization problem, however they also allow for the imposition of soft constraints. As such, the constraint bounds themselves should be be chosen conservatively to ensure adherence to physical constraints. This formulation helps maintain the balance between constraint violation and control feasibility ensuring performance even under challenging dynamic conditions.

\section{Experimental Setup}\label{sec: Experimental Setup}
This section describes the control scheme being tested and the experimental structure used to assess it. A summary of the controller, the testbed used for experimentation, and the RTA formulation can be found above in the \textit{Background}. Details regarding experimental methodology and structure is discussed first. This is followed by training details, factors for evaluation, and performance criteria. 

\subsection{Methodology}
The purpose of our experiments is to assess some of the practical consequences of using DRL based motion control while operating under high-risk conditions in close proximity of other space vehicles. Our experiments are designed around the specification of multiagent encounters that are designed to result in a high-likelihood of interagent collision. In practice, effective HL planners will not assign tasks to LL planners resulting in such conditions. This is affirmed by the results of \cite{Lei25}, where RTA intervention was present but relegated primarily to speed correction induced by hardware-in-the-loop sensor feedback. In this capacity, both the HL and LL policies were able to effectively continue operation. Positional correction was not a primary intervention factor as the HL policy was effective at sending safe waypoints to the LL controller. To help assess scenarios where this safety is not guaranteed implicitly by policy specification itself, we instantiate agents with the following considerations:
\begin{itemize}
    \item[1.] The HL planner outputs predetermined waypoints to the DRL based LL control policy. These are specified for a \textit{Three Agent Standoff}, including two deputy agents and a chief object specified at frame center. Agent 1 starts at $\delta \mathbf{r}_{1}(0) = (-200\text{m},0,0)$ with alternating waypoint goals specified by: $$\delta \mathbf{r}_{g}^{(1)} = (300\text{m},0,0) \text{ and } \delta \mathbf{r}_{g}^{(2)} = -\delta \mathbf{r}_{g}^{(1)}. $$ Agent 2 starts at $\delta \mathbf{r}_{2}(0) = (0,-200\text{m},0)$ with alternating waypoint goals specified by:
    $$\delta \mathbf{r}_{g}^{(3)}  = (0,300\text{m},0) \text{ and } \delta \mathbf{r}_{g}^{(4)}  = -\delta \mathbf{r}_{g}^{(3)}. $$
    For an example of this configuration, please see \cref{fig:expr_2} below.
    \item[2.] The LL control policy for agent-$i$ only has access to its own state information and cannot observe other vehicles in the environment. To support this, the training environment from \cite{Lei22} is modified to reflect single agent operation rather that multiagent. Specifically, observations are formed according to \cref{eqn:llobs} and the reward function is of the form \cref{eqn:rew_ll}.
\end{itemize}
The training environment and reward formulation associated with LL control imposed in consideration [2] is designed to incentivize distance minimization according to the Euclidean norm. With this in mind, nominal reference trajectories can be easily constructed satisfying all relevant velocity constraints. Movement between waypoints is optimally symmetric with respect to reflection across the origin. Under ideal operation, the waypoint goal specification imposed in a three agent standoff guarantees collision between: Agent 1 \& Agent 2, Agent 1 \& Chief, and Agent 2 \& Chief. It should be noted that this is defined to ensure feasible control inputs exist when RTA intervenes; for additional details, please refer to the Background section on RTA. Meanwhile, consideration [2] requires the training of \textit{single agent} policies to ensure that the impact of RTA intervention is reducible to the structure of the motion control policy itself (eg. a neural network) instead of being implicitly biased through compounding multi-agent safety considerations. 

\subsection{Factors for Evaluation}

When conducting tests in the LINCS lab, it is important to carefully structure experiments to ensure traceability in controller performance change. This amounts to leveraging both simulation and quadcopter emulation capabilities. Among other things, the LINCS simulation environment allows for testing of the following key factors:
\begin{itemize}
    \item Timing asynchronicity in the control loop for policy inference;
    \item High-fidelity environmental dynamics including the J2 perturbation;
    \item Real-time RTA for configurable control intervention and safety enforcement.
\end{itemize}

The LINCS Lab also supports mixed setups, where agents operate in both simulation and emulation environments simultaneously. For example, in a scenario with \(n=3\) agents, two agents may operate in simulation while the third is emulated via a quadcopter. This mixed setup enables testing of system scalability and robustness through fine-tuned experimental control over a variety of test factors including: 
\begin{itemize}
    \item All LINCS simulation factors;
    \item HIL quadcopter state feedback.
\end{itemize}

We leverage this by conducting three distinct sets of experiments under gradually increasing complexity. The first set is used to evaluate the effect of timing asynchronicity and dynamics perturbation on single agent point to point motion control tasks. Timing asynchronicity describes the difference between environmental step rate used by LINCS and the environmental step rate used during training of the LL control policy. This is accompanied by single agent trials conducted in the LINCS emulation environment. Side-by-side with LINCS simulation, the only experimental factor being assesed at this step is the inclusion of real-time quadcopter feedback to the control loop. Concurrently tested for hierarchical control in \cite{Lei25}, it is expected that performance degrades but task completion remains stable.

The second set of experiments builds from the first through the utilization of a mixed emulation setup used to instantiate the three agent standoff scenario described above. Although LINCS has the capability of operating multiple physical agents in the same trial, the mixed setup helps remove quadcopter-to-quadcopter interaction effects. More specifically, it ensures that there cannot be a collision event between quadcopters and removes the potential impact of propwash on peer agent dynamics. On the policy level, since each agent cannot observe other agents' behavior, results are expected to be consistent with the outcome of the first experimental suite. This is important because it allows for the establishment of an experimental baseline for evaluating multiagent operation in the absence of direct RTA intervention.

The final set of experiments uses the same three agent standoff but activates RTA. This will result in direct control intervention for collision avoidance between both agents and the chief. In conjunction with the first and second suite of experiments, this allows us to independently analyze the implications of each test factor. Performance is measured across all experiments through the following:
\begin{itemize}
    \item Targets Reached: This counts the number of waypoint targets that are successfully reached by the LL controller without instance time out. A single point to point LL control task cannot exceed $500\text{s}$.
    \item Time Taken: The total time recorded during an experiment.
    \item Distance Traveled: The sum of distance traveled across all agents during an experiment.
    \item Fuel Consumption: The sum of fuel consumption across all agents during an experiment. It should be noted, that this metric is used to evaluate relative changes - not absolute. The underyling LL control policy is not designed to minimize fuel, it instead minimizes distance. Keeping track of this helps describe ancillary effects the test factors may have on performance.
\end{itemize}
For a brief summary of each experiment, please refer to \cref{tab:experiment_outline} below.

\begin{table}[h!]
    \centering
    \resizebox{\textwidth}{!}{%
    \begin{tabular}{|c|c|p{8cm}|}
        \hline
        \textbf{Experiment} & \textbf{Configuration} & \textbf{Purpose} \\ \hline
        Experiment 1 & Single Agent, Single Chief, No RTA & Evaluate baseline performance of LL-controller acting against perturbation effects, timing asynchronicity, and HIL noise.\\ \hline
        Experiment 2 & Three Agent Standoff, No RTA & Confirm consistency with Experiment 1 results under consideration [1] and [2].\\ \hline
        Experiment 3 & Three Agent Standoff, RTA & Evaluate performance of LL-controller when faced against significant RTA intervention effects. \\ \hline
    \end{tabular}%
    }
    \caption{Summary of experiments used for evaluating performance degradation of the LL controller from \cite{Lei22} when exposed to challenging operational circumstances.}
    \label{tab:experiment_outline}
\end{table}

\subsection{Parameter Specification}
Key parameters used in the LINCS Lab experiments, specifically for RTA and LL control policy reward coefficients are detailed below in \cref{tab:rta_parameters}. 

\begin{table}[h!]
    \centering
    \begin{tabular}{|c|c|c|c|}
        \hline
        \textbf{Parameter} & \textbf{Value} & \textbf{Coeff.} & \textbf{Value}\\ \hline
        Collision Radius (\(r_c\)) & 50 m & $\alpha$ & $1\times 10^{-3}$\\ \hline
        Maximum Velocity (\(v_c\)) & 3 m/s & $\beta$ & $1\times 10^{-2}$ \\ \hline
        Maximum Acceleration (\(a_c\)) & 1.732 m/s\(^2\) & $\nu$ & $1\times 10^{-2}$\\ \hline
        Input Upper Bound (\(f_c\)) & 1 N & $\sigma_{\mu}$ & $3.08\times 10^{-1}$\\ \hline
        Acceptance (LINCS) & 15m & Acceptance (Training) & 10m\\ \hline
    \end{tabular}
    \caption{RTA Parameters and reward coefficients used for LINCS Experiments.}
    \label{tab:rta_parameters}
\end{table}
\vspace{-10pt}
The choices for these are motivated largely by the work in \cite{Lei22} and \cite{Lei25}. Training implementation was specified to ensure stability and consistency with \cite{Lei22}. RTA and lINCS parameter specification was chosen with the following rationale:
\begin{itemize}
    \item The collision radius of 50m was set to ensure that the positional constraint of the RTA would have a high likelihood ($>95\%$) under the range of nominal sampling trajectories experienced in the LL controller's training environment. 
    \item The velocity constraint was chosen for consistency with \cite{Lei25}, which set an aggresive threshold for the testing of RTA intervention focused on the violation of speed limits along reference trajetories.
    \item Acceleration limits are set specific to vehicle mass for consistent output that is binding when the input force is maximized (eg. achieves the input upper bound for each thruster).
    \item The input upper bound is set to the same value as used during training the LL controller - as specified in \cite{Lei22}.
    \item The acceptance threshold for a success condition to be triggered at waypoint arrival is set more aggressively in training versus rollout. This helps balance the priority between velocity stabilization and subsequent task assignment. 
\end{itemize}

\section{Results}\label{sec: Results}
This section outlines key results for each experiment conducted. These are organized by the hierarchy presented in \cref{tab:experiment_outline}.  

\FloatBarrier
\subsection{Experiment 1: Single-Agent Motion Control}

The first set of experiments aims to characterize the baseline performance of the DRL single agent controller across the three different deployment environments. The performance in the training environment is recorded in \cref{tab:SA_baseline} and a representative trajectory is displayed in \cref{fig:LL Controller}. It is important to note that the numbers in \cref{tab:SA_baseline} represent an extrapolated baseline for scenario-based performance rather than a specific rollout of the scenario considered. This is to help combat sampling bias in the single instance episodes conducted in LINCS. Our baseline considers a random sample of 50 trials with each trial conducted in the training environment using \cref{eqn:sampling} to generate randomized starting positions and goal waypoints. Under these conditions, the learned policy is able to move to the assigned waypoint with a 100\% success rate. On average, the trained agent takes $168.82\text{s} (SD=59.95\text{s})$ to complete its task, and travels an average of $622.27 \text{m} (SD=303.72\text{s})$ meters. With an optimal translational trajectory represented by a straight line between the agent's initial position and its target waypoint, the sampled trajectories are within $72.69\text{m} (SD=82.13\text{m})$ when measured by total distance traveled, a 13.2\% increase. Using this as an environmental baseline, we normalize performance metrics by a fixed single unit of time and distance measured against the straight-line optimal value. The LL controller takes $0.307\text{s}$ (on average) to cover a single meter while moving a total of $1.132\text{m}$ relative to an optimal straight-line distance of $1\text{m}$.

For consistency with the three-agent standoff configuration, we conduct two LL control trials in LINCS lab where the agent is tasked to repeatedly maneuver between target locations, starting from an initial location at $(-200\text{m}, 0, 0)$ and making two back and forth passes between $(\pm 300\text{m}, 0, 0)$. The single-unit performance metrics are used to estimate idealized values for this scenario through a multiplicative factor of idealized (total) straight-line distance of 2300m. Fuel cost is estimated in a similar way - it is important to recall that this is an ancillary metric that does not (in isolation) reflect the performance of the trained LL controller; for final values refer to \cref{tab:SA_baseline}.

\begin{table}[htb!]
    \centering
    \resizebox{0.6\textwidth}{!}{%
    \begin{tabular}{|c|c|c|c|}
        \hline
        \textbf{Performance Metric} & \textbf{Training Baseline} & \textbf{LINCS SIM} & \textbf{LINCS CPS}\\
        \hline
        Targets Reached & 4 & 4 & 4\\
        \hline
        Time Taken (s) & 706.64 & 709.0 & 1109.65\\ \hline
        Distance Traveled (m) & 2604.68 & 2639.55& 4994.38\\ \hline
        Fuel Consumption (\(\Delta V\)) (m/s) & 82.59 & 81.83 & 475.35\\
        \hline
    \end{tabular}
    }
    \caption{Experiment 1 summary statistics}
    \label{tab:SA_baseline}
\end{table}

\begin{figure}[!htb]
\begin{subfigure}[b]{0.5\textwidth}
    \includegraphics[width=\textwidth]{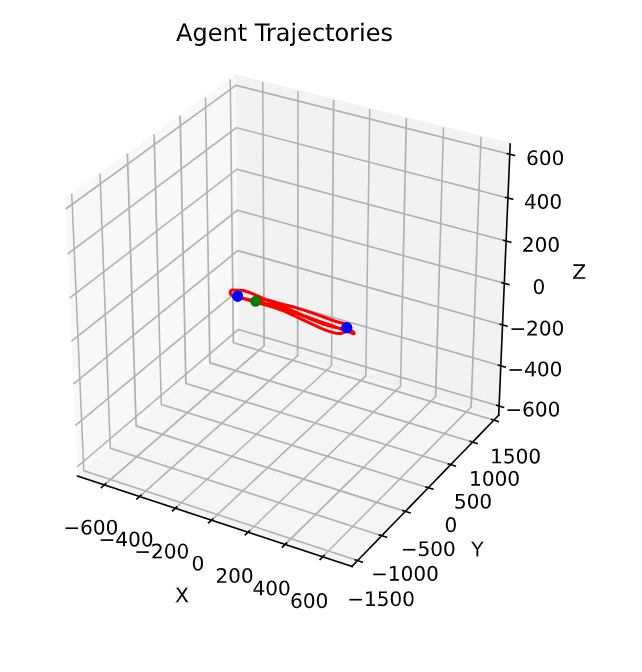}
\end{subfigure}
\hfill
\begin{subfigure}[b]{0.5\textwidth}
    \includegraphics[width=\textwidth]{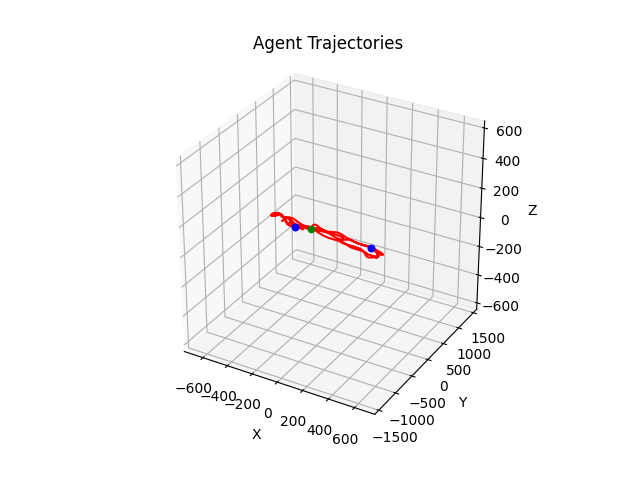}
\end{subfigure}
\hfill
\begin{subfigure}[b]{0.45\textwidth}
    \includegraphics[width=\textwidth]{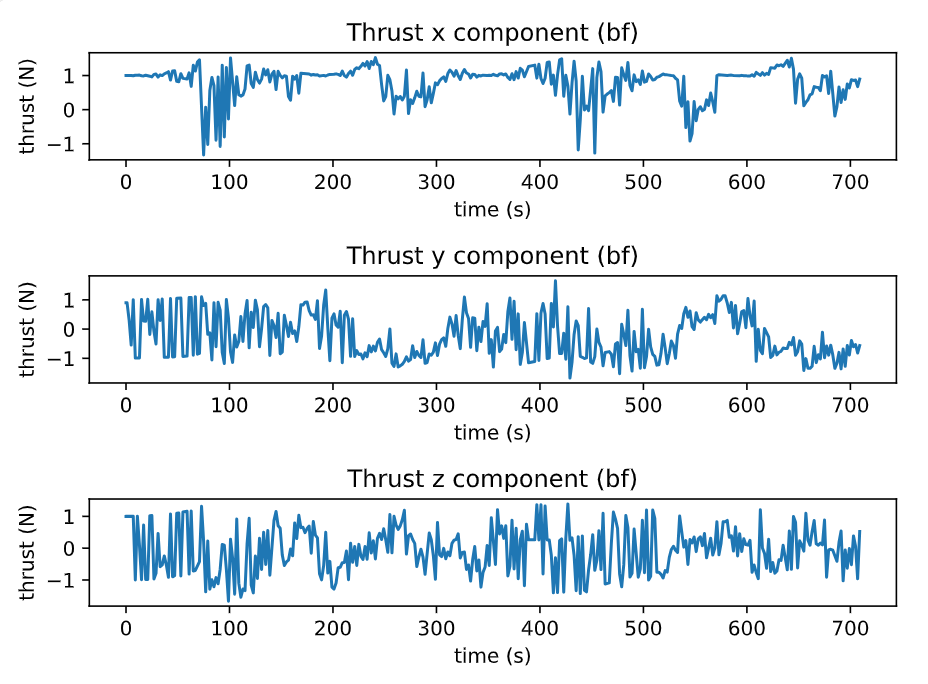}
\end{subfigure}
\hfill
\begin{subfigure}[b]{0.45\textwidth}
    \includegraphics[width=\textwidth]{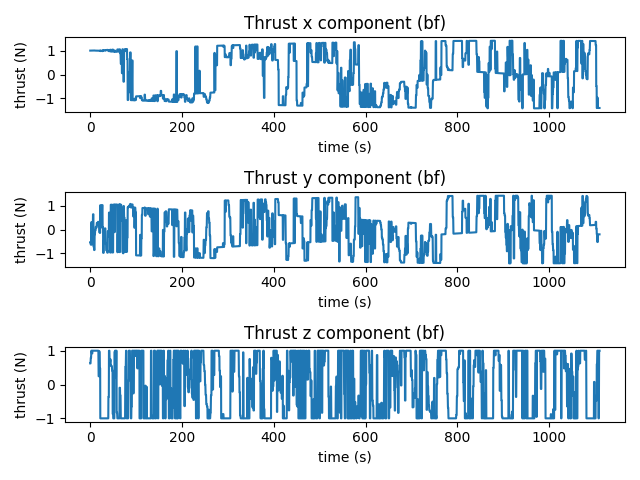}
\end{subfigure}
\caption{Experiment 1 test trajectories. The left hand panel shows LINCS SIM results, the right hand panel show LINCS CPS results.}
\label{fig:expr_1}
\end{figure}

When tested in the LINCS simulation environment (LINCS SIM), the LL control policy performs within error to the training baseline with only a $0.33\%$, $1.34\%$ and $-0.91\%$ difference respectively. Qualitatively, these results are consistent with our expectations. There are two primary test factors being considered: timing asynchronicity and higher fidelity environmental dynamics (including J2). More specifically, the training environment assumes that the agents are operating using linearized relative motion dynamics, while the LINCS simulation engine uses nonlinear two body dynamics with a J2 perturbation. In combination with the close spatial proximity (to the RSO) the linearization of the nonlinear dynamics is relatively accurate; see \cite{Clohessy60}. Additionally, the J2 perturbation is applied equally to both parties in the relative motion scheme and the time horizon for this mission is far shorter than that which the J2 perturbation typically becomes relevant; see \cite{Mishne04}. More details can be found in \cref{tab:SA_baseline}; the LINCS SIM single agent trajectory can be seen in \cref{fig:expr_1}.

Building on the LINCS SIM results, we test the same LL control policy in the LINCS cyber-physical systems testbed (LINCS CPS) with a quadcopter emulating spacecraft dynamics in the control loop. Immediately, we note severe performance degradation relative to the two previous environments. The emulated agent takes $57.03\%$ longer to complete the 4 waypoint trial and travels $91.59\%$ farther to do so (relative to the LINCS SIM trial). The impact of this is especially sensitive in fuel utilization where the measurement delta was $475.59\%$ for the HIL agent. Qualitatively, we note a couple of errant behaviors:
\begin{enumerate}
    \item The HIL agent exhibits minor oscillatory behavior during the course of a point-to-point trajectory, likely due to the overshooting and rebound effects induced by the quadrotor's trajectory tracking controller combined with perturbations induced by atmospheric effects.
    \item The HIL agent struggles to converge to its target location, and circles around the goal position until it inches close enough to trigger the termination condition.
\end{enumerate}  

Though the waypoint acceptance region was increased from $10\text{m}$ to $15\text{m}$ between the training environment and the LINCS environments, the physical agent demonstrates circling behavior at a much higher intensity than in any of the simulated agents. Despite this, all waypoint targets were reached without hitting the timeout threshold of 500s per waypoint task. When comparing LINCS SIM and CPS results in \cref{fig:expr_1}, the additional trajectory noise and stalling behavior is evident. Additionally, the potential impact of sensor noise on LL-control can be seen when comparing the thrust profile of the simulated and emulated agent. All else equal, the HIL agent is attempting to provide full thrust in the z-direction much more frequently than its simulated counterpart. 

\FloatBarrier
\subsection{Experiment 2: Multi-Agent Baseline}
\begin{table}[h!]
    \centering
    \resizebox{0.6\textwidth}{!}{%
    \begin{tabular}{|c|c|c|}
        \hline
        \textbf{Performance Metric} & \textbf{LINCS SIM} & \textbf{LINCS CPS}\\
        \hline
        Targets Reached & 8 & 8\\
        \hline
        Time Taken (s) & 745.0 & 1431.83\\ 
        \hline
        Distance Traveled (m) & 5359.42& 8946.41\\ 
        \hline
        Fuel Consumption (\(\Delta V\)) (m/s) & 167.89 & 874.61\\
        \hline
    \end{tabular}
    }
    \caption{Experiment 2 summary statistics.}
    \label{tab:MA_expr_no_rta}
\end{table}

\begin{figure}[!htb]
\begin{subfigure}[b]{0.45\textwidth}
    \includegraphics[width=\textwidth]{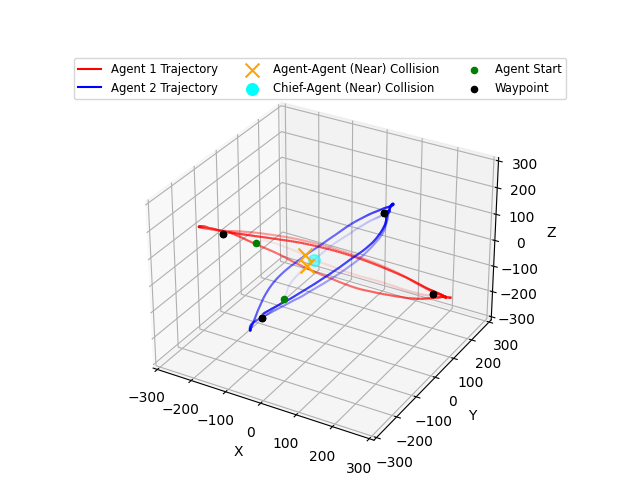}
\end{subfigure}
\hfill
\begin{subfigure}[b]{0.45\textwidth}
    \includegraphics[width=\textwidth]{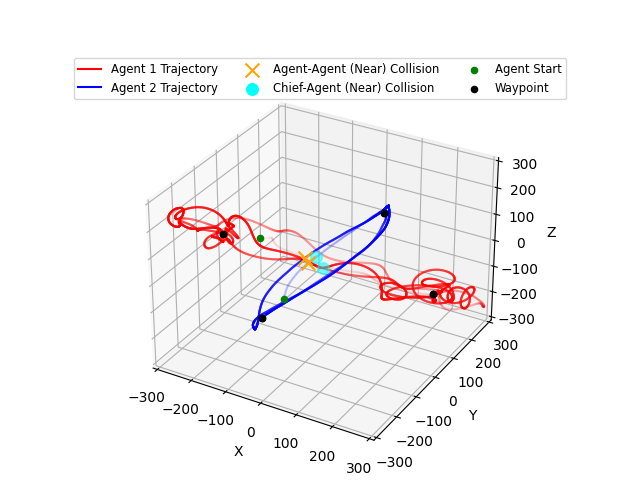}
\end{subfigure}
\hfill
\begin{subfigure}[b]{0.45\textwidth}
    \includegraphics[width=\textwidth]{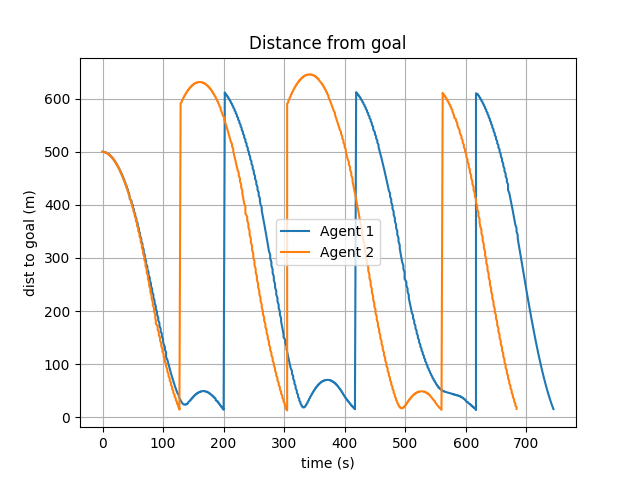}
\end{subfigure}
\hfill
\begin{subfigure}[b]{0.45\textwidth}
    \includegraphics[width=\textwidth]{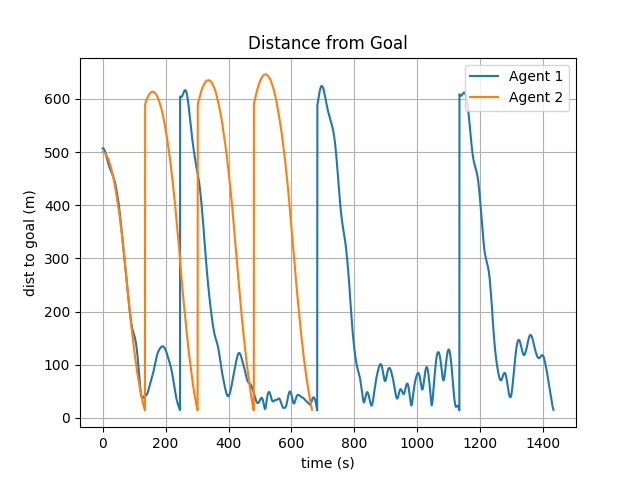}
\end{subfigure}
\hfill
\begin{subfigure}[b]{0.5\textwidth}
    \includegraphics[width=\textwidth]{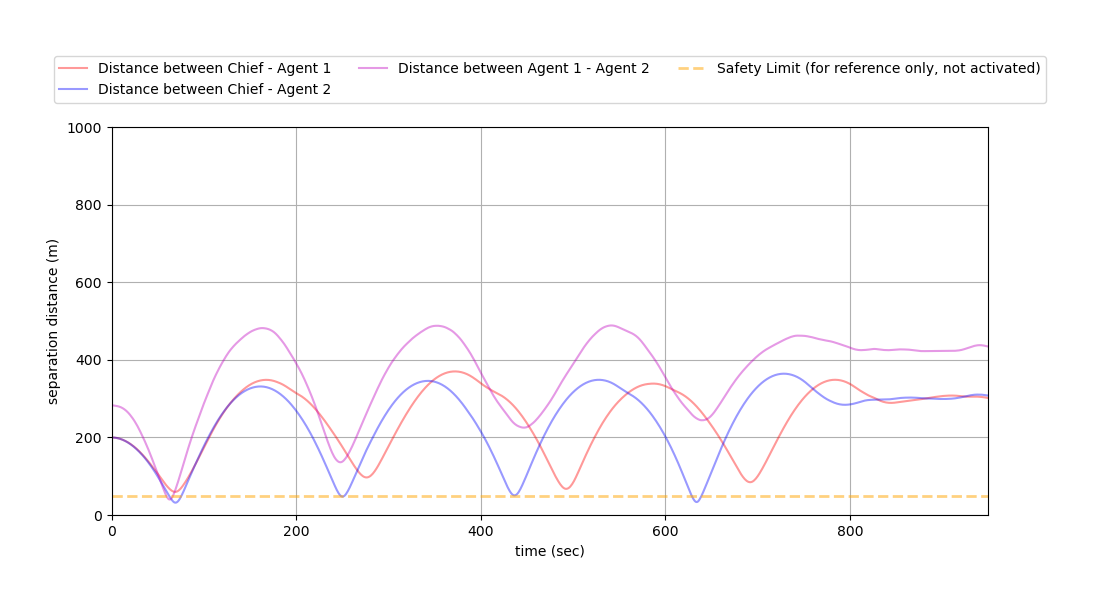}
\end{subfigure}
\hfill
\begin{subfigure}[b]{0.5\textwidth}
    \includegraphics[width=\textwidth]{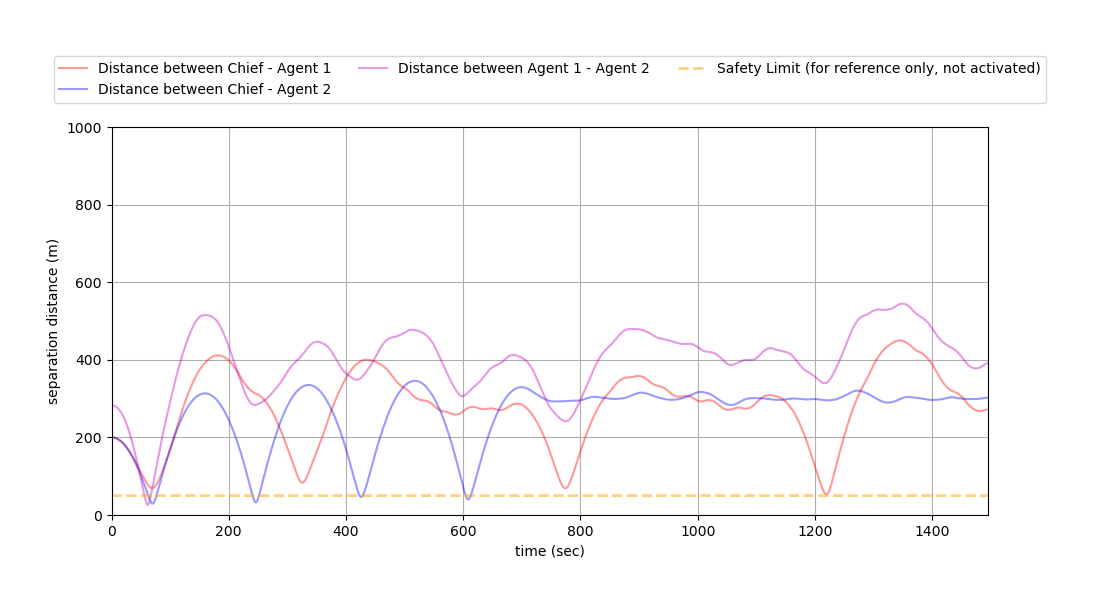}
\end{subfigure}
\hfill
\caption{Experiment 2 test trajectories. The left hand panel shows LINCS SIM results, the right hand panel show LINCS CPS results.}
\label{fig:expr_2}
\end{figure}

The second set of experiments instantiates the three-agent standoff scenario described in \textit{Experimental Setup} above without external RTA in the control loop. The two controllable agents are homogeneous and share separate copies of the same LL control policy. Because the observation space and reward function used for training excludes multiagent information, there should be no interaction effects present in their trajectories while RTA is deactivated. Indeed, comparing the trajectory of agent 1 (LINCS SIM) with that in the single agent experiment (LINCS SIM), we see that there is only a $4.30\%$, $5.22\%$, and $7.33\%$ in the time, distance and fuel metrics respectively. Meanwhile, the second agent (LINCS SIM) acting on a cross path shows a $-8.04\%$, $-5.35\%$, and a $-8.84\%$ difference in time, distance and fuel respectively when compared against the single-agent (LINCS SIM). Given the expectations for variance established in the single-agent training baseline (see \cref{tab:SA_baseline}), this disparity is quite small. Looking at \cref{fig:expr_2}, the crossing times defined by local minima in interagent distance are consistent with respect to arrival time for agent-chief interaction. High-level observation indicates a gradual drift in arrival time for agent-agent crossing, perhaps due to the aggregation of minor timing effects that can occur between the two distinct sets of waypoint tasks assigned to agent 1 and agent 2. 

Next, a three agent standoff trial was conducted in the LINCS CPS testbed. This included a hybridized setup with 2 simulated agents (1 controllable, 1 RSO) and 1 emulated agent (controllable). For consistency with Experiment 1, the emulated agent is assigned to the waypoints defined above as: $\delta \mathbf{r}_{g}^{(1)}$ and $\delta \mathbf{r}_{g}^{(2)}$. In this trial, the emulated agent retains both the oscillatory and circling behaviors evident in the first experiment, taking a longer and less regular trajectory than in the simulated results. Summarily, the trajectory of agent 1 and the emulated \textit{single agent} exhibit similar performance characteristics. Ceteris paribus, the trajectory of agent 2 in the hybrid setup and the simulated agent 2 in the simulated setup are directly comparable. This supports the hypothesis there isn't a direct interaction effect between agents. If so, the significant performance degradation of agent 1 would likely have impacted the performance of agent 2; although more testing is needed to formally confirm. Although, it doesn't appear that there is a discernible interaction effect impacting agent behavior, there is a significant impact the performance delta of Agent 1 has on crossing times. In this setting, though the pair of agents have one crossing in the beginning pass, the timings quickly separate, and there are no regular interactions beyond. This can be seen reflected in the ``Distance from goal" plots in \cref{fig:expr_2}. 

\FloatBarrier
\subsection{Experiment 3: RTA Intervention}

\begin{table}[h!]
    \centering
    \resizebox{0.6\textwidth}{!}{
    \begin{tabular}{|c|c|c|}
        \hline
        \textbf{Performance Metric}  & \textbf{LINCS SIM} & \textbf{LINCS CPS}\\
        \hline
        Targets Reached & 8 & 8\\
        \hline
        Time Taken (s) & 1296.0 & 3418.32\\
        \hline
        Distance Traveled (m) & 5167.28 & 18822.40\\ 
        \hline
        Fuel Consumption (\(\Delta V\)) (m/s) & 240.16 & 2025.92\\
        \hline
    \end{tabular}
    }
    \caption{Experiment 3 summary statistics.}
    \label{tab:MA_expr_rta}
\end{table}

Similar to the second set of experiments, Experiment 3 is designed as a three agent standoff, with an additional RTA optimization controller enabled in the LINCS control stack. Details regarding its formulation can be seen above in \textit{Background}, and relevant parameter specifications can be found in \cref{tab:rta_parameters}.

\begin{figure}[!htb]
\begin{subfigure}[b]{0.45\textwidth}
    \includegraphics[width=\textwidth]{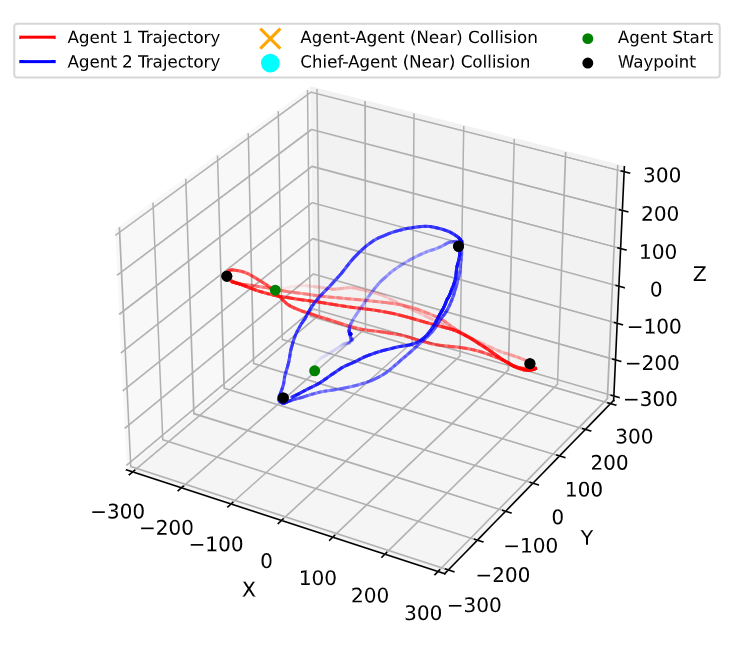}
\end{subfigure}
\hfill
\begin{subfigure}[b]{0.45\textwidth}
    \includegraphics[width=\textwidth]{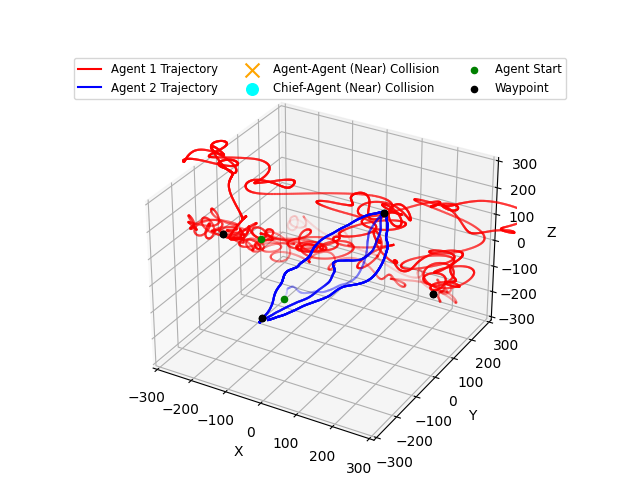}
\end{subfigure}
\hfill
\begin{subfigure}[b]{0.45\textwidth}
    \includegraphics[width=\textwidth]{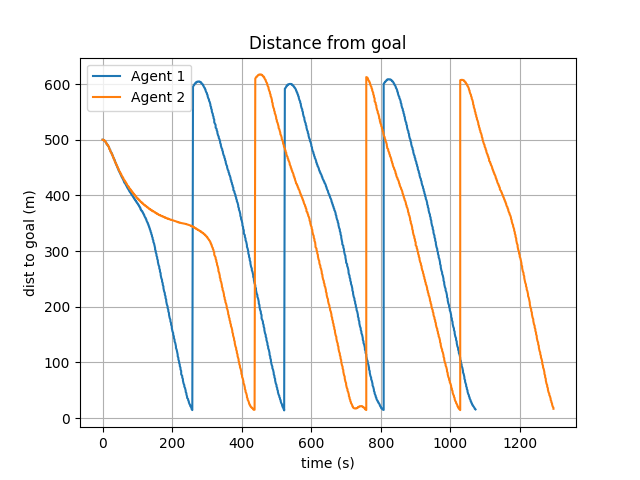}
\end{subfigure}
\hfill
\begin{subfigure}[b]{0.45\textwidth}
    \includegraphics[width=\textwidth]{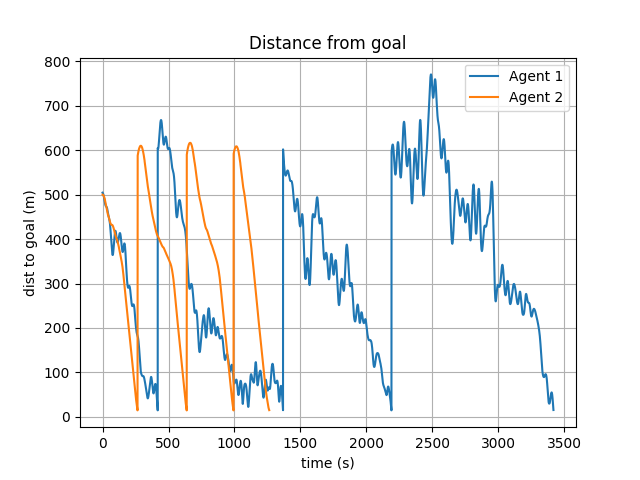}
\end{subfigure}
\caption{Experiment 3 test trajectories. The left hand panel shows LINCS SIM results, the right hand panel show LINCS CPS results.}
\label{fig:expr_3_HL}
\end{figure}

\begin{figure}[!htb]
\begin{subfigure}[b]{0.45\textwidth}
    \includegraphics[width=\textwidth]{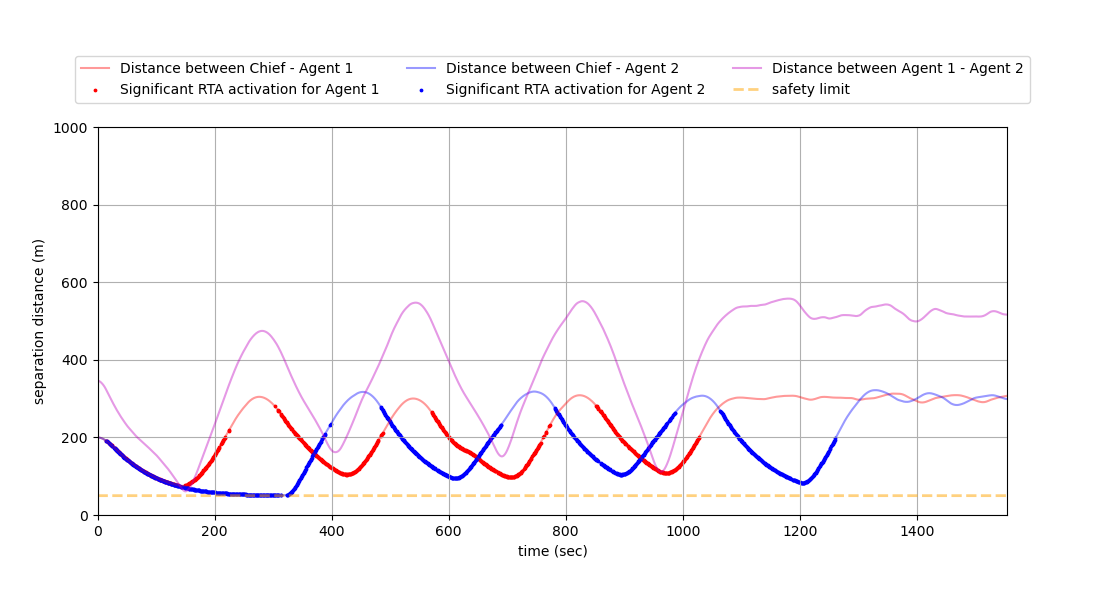}
\end{subfigure}
\hfill
\begin{subfigure}[b]{0.45\textwidth}
    \includegraphics[width=\textwidth]{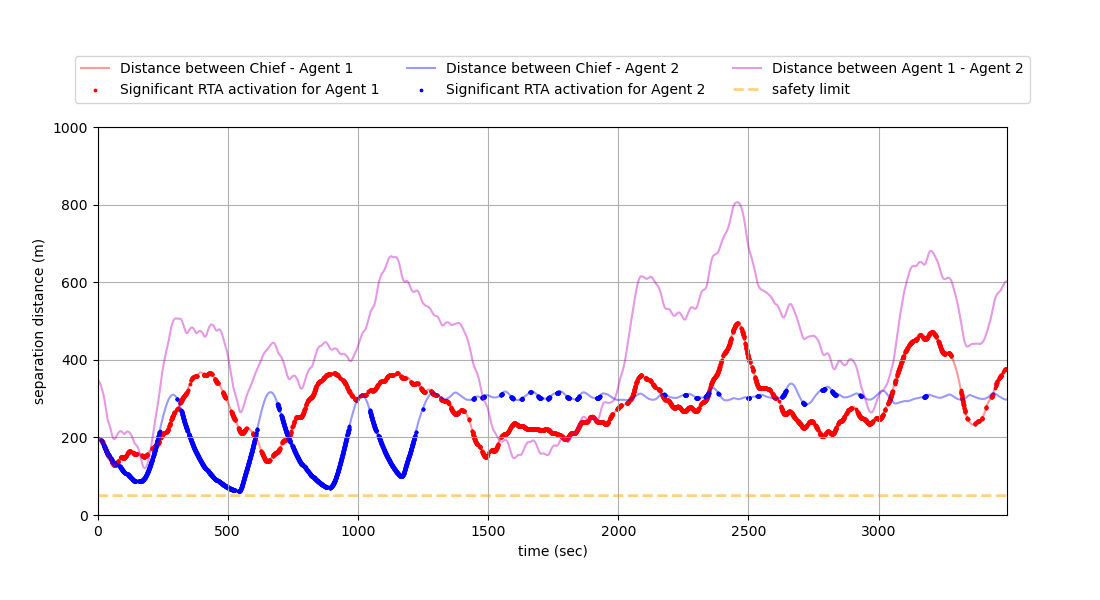}
\end{subfigure}
\hfill
\begin{subfigure}[b]{0.45\textwidth}
    \includegraphics[width=\textwidth]{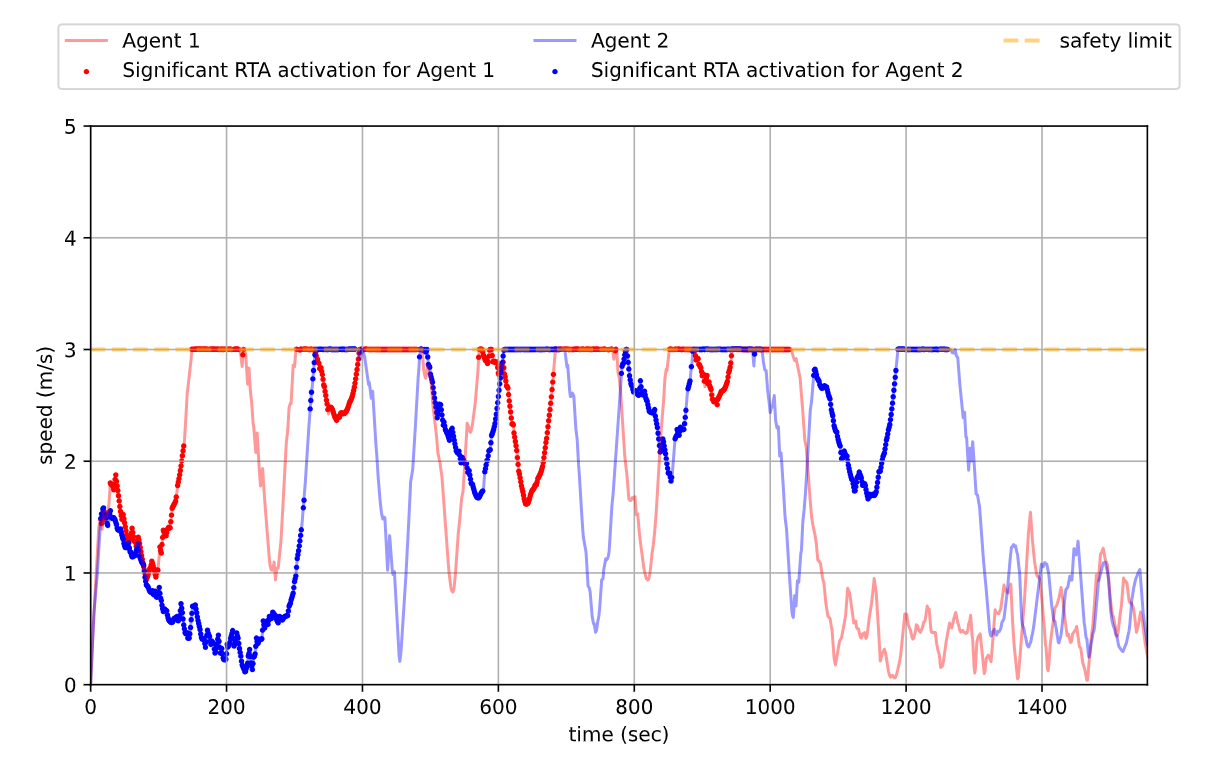}
\end{subfigure}
\hfill
\begin{subfigure}[b]{0.45\textwidth}
    \includegraphics[width=\textwidth]{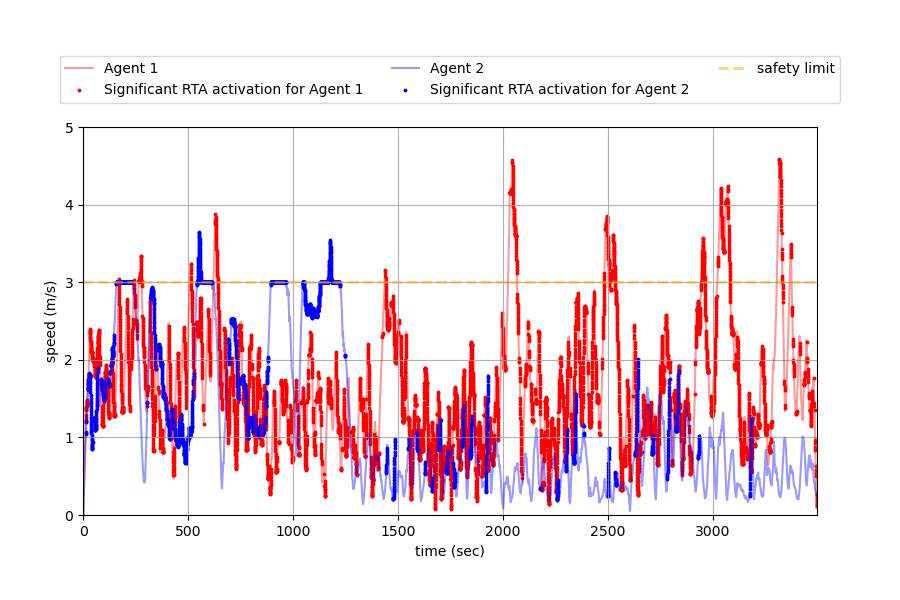}
\end{subfigure}
\caption{Experiment 3 RTA activation. The left hand panel shows LINCS SIM results, the right hand panel show LINCS CPS results.}
\label{fig:expr_3_RTA}
\end{figure}

At a glance, the pair of DRL trained LL controllers are able to reach all eight of their target waypoints throughout the course of the experiment - in both LINCS SIM and LINCS CPS. In LINCS SIM, the agents take a total of $1296.0\text{s}$ to complete the experiment, cumulatively traveling $5167.28\text{m}$ and expending $240.16\text{m}/\text{s}$ of delta V. Compared to the equivalent case in Experiment 2, these represent a $+73.96\%$, $-3.59 \%$ and a $+43.05 \%$ change respectively. We observe that the increase in cumulative distance is comparatively much smaller than the increase in total time taken, which is an effect of the RTA intervention limiting agent speed. Accordingly, we can see that the maximum speed of the agents in \cref{fig:expr_3_RTA} reaches about $7 \text{m}/\text{s}$ at the apex of the trajectory in the nominal case, but is capped at $3\text{m}/\text{s}$ under RTA intervention; see \cref{fig:expr_3_RTA}. 

When compared with Experiment 2, the test conducted in LINCS SIM saw significant control intervention by the RTA controller acting on behalf of both agents. This was prompted largely from persistent operation on the boundary of the imposed interagent distance and speed limit; see \cref{fig:expr_3_HL} and \cref{fig:expr_3_RTA}. The RTA was consistently active when nearing agent-agent and agent-chief crossings. This provides challenging circumstance as it generally correlates with the times in which each agent is traveling the most quickly. At the test start, there is an extended period of RTA activation, as the agents are near their initial passes with the chief, but successfully maneuver to avoid it. Afterwards, the agents are no longer in alignment, with one operating within a vertical plane, with the other agent orthogonal to it. The displaced trajectories with the RTA enforced speed limit take significantly longer finish. Although the RTA significantly intervenes in the control loop, the agents are still able to complete their tasks without timing out.

Within the LINCS CPS test environment, we immediately notice destabilization of the emulated agent (Agent 1). Oscillatory and circling behaviors are more pronounced in this environment-- presumably due to the interaction between the VICON sensor feedback, physical quadrotor effects, and the frequent RTA intervention-- leading to vastly less efficient trajectories; see the left vs right panel in \cref{fig:expr_3_HL}. Comparing the performance metrics in \cref{tab:MA_expr_rta} to the equivalent LINCS CPS analogue in Experiment 2 (see \cref{tab:MA_expr_no_rta}), the presence of RTA activation impacted mission duration by over $138.74\%$, cumulative distance traveled by $110.39\%$, and fuel utilization by $131.64\%$.

We can see from \cref{fig:expr_3_RTA}, that the RTA is persistently active, though there are few close collision events, indicating that it is primarily trying to prevent the physical agent from violating the velocity constraints. The RTA is largely able to do so, but there are instances where agent velocity spikes over what the RTA overseer can effectively control. Intuitively, the RTA is architected as a model-based optimization controller, and while the emulated agent attempts to track trajectories in the space frame as accurately as possible, there remain unmodeled artifacts induced by atmospheric effects and imprecise tracking controls that cause the constraint violations. In turn, the RL controller and RTA stack also struggle to perform the smooth and performant trajectories of the other experiments.

\FloatBarrier
\section{Conclusions}\label{sec: Conclusion}
This study evaluated the performance and robustness of a DRL based LL motion controller for close proximity satellite operation when placed under challenging, low-probability scenarios. Complementing the studies in \cite{Lei25}, key challenges in autonomous satellite operations, including task allocation, trajectory optimization, and adaptation to dynamic uncertainties are experimentally investigated. Leveraging LINCS testing facilities, we placed the LL controller in increasingly difficult situations and identified the specific test factors contributing to significantly degraded performance. In combination with the positive results in \cite{Lei25}, the LINCS CPS experiments underscored the impact that physical dynamics, feedback delays, and sensor noise can have on controller performance. In real-world conditions when faced with \textit{infrequent} RTA intervention, the controller can effectively perform its guidance tasks. However, when faced with challenging conditions - specifically scenarios where inter-agent collision is highly likely, the LL controller destabilizes and cannot effectively complete its tasking. This study helps highlight areas that would benefit from future study and more robust statistical analysis. It is a prerequisite to enabling robust and efficient autonomous satellite operations in increasingly complex and dynamic environments.

\bibliography{references}
\end{document}